\documentclass[sigconf]{acmart}

\usepackage{booktabs} 
\usepackage{graphicx}
\usepackage{subfigure}

\usepackage{algorithm}
\usepackage{algorithmic}
\usepackage{multirow}
\usepackage{array}
\usepackage{color}
\usepackage{amsmath,amsfonts,amssymb,bbm,epsfig,bm}
\usepackage{float}
\usepackage{enumerate}
\usepackage{balance}

\usepackage{appendix}

\setcopyright{rightsretained}

\acmDOI{10.475/123_4}

\acmISBN{123-4567-24-567/08/06}

\acmConference[KDD'17]{ACM KDD conference}{August 2017}{Halifax, Nova Scotia, Canada} 
\acmYear{2017}
\copyrightyear{2017}

\acmPrice{15.00}

\newcommand{\E}{\mathbb{E}}

\newcommand{\eqtref}[1]{Eq.\eqref{#1}}

\makeatletter
\def\BState{\State\hskip-\ALG@thistlm}
\makeatother

\begin{document}
\title{Efficient Correlated Topic Modeling with Topic Embedding}

\author{
Junxian He$^{*1,3}$,~ Zhiting Hu$^{*1,2}$,~ Taylor Berg-Kirkpatrick$^1$,~ Ying Huang$^{3}$,~ Eric P. Xing$^{1,2}$ \\
{\large 
Carnegie Mellon University$^{1}$ 
\quad Petuum Inc.$^2$ 
\quad Shanghai Jiao Tong University$^3$
}\\
{\large $^{*}$ Equal contribution\quad \{junxianh,zhitingh,tberg,epxing\}@cs.cmu.edu
,~ hy941001@sjtu.edu.cn}
}

\copyrightyear{2017} 
\acmYear{2017} 
\setcopyright{acmcopyright}
\acmConference{KDD'17}{}{August 13--17, 2017, Halifax, NS, Canada.}
\acmPrice{15.00}
\acmDOI{10.1145/3097983.3098074}
\acmISBN{978-1-4503-4887-4/17/08} 

\begin{abstract}
Correlated topic modeling has been limited to small model and problem sizes due to their high computational cost and poor scaling. In this paper, we propose a new model which learns compact topic embeddings and captures topic correlations through the closeness between the topic vectors. Our method enables efficient inference in the low-dimensional embedding space, reducing previous cubic or quadratic time complexity to {\it linear} w.r.t the topic size. We further speedup variational inference with a fast sampler to exploit sparsity of topic occurrence. Extensive experiments show that our approach is capable of handling model and data scales which are several orders of magnitude larger than existing correlation results, without sacrificing modeling quality by providing competitive or superior performance in document classification and retrieval.
\end{abstract}



\maketitle

\section{Introduction}
Large ever-growing document collections provide great opportunities, and pose compelling challenges, to infer rich semantic structures underlying the data for data management and utilization.
Topic models, particularly the Latent Dirichlet Allocation (LDA) model~\cite{blei2003latent}, have been one of the most popular statistical frameworks to identify latent semantics from text corpora. One drawback of LDA derives from the conjugate Dirichlet prior, as it models topic occurrence (almost) independently and fails to capture rich topical correlations (e.g., a document about virus may be likely to also be about disease while unlikely to also be about finance). Effective modeling of the pervasive correlation patterns is essential for structural topic navigation, improved document representation, and accurate prediction~\cite{ranganath2016correlated,blei2007correlated,chen2013scalable}. Correlated Topic Model (CTM)~\cite{blei2007correlated} extends LDA using a logistic-normal prior which explicitly models correlation patterns with a Gaussian covariance matrix.

Despite the enhanced expressiveness and resulting richer representations, practical applications of correlated topic modeling have unfortunately been limited due to high model complexity and poor scaling on large data. For instance, in CTM, direct modeling of pairwise correlations and the non-conjugacy of logistic-normal prior impose inference complexity of $\mathcal{O}(K^3)$, where $K$ is the number of latent topics, significantly more demanding compared to LDA which scales only linearly. While there has been recent work on improved modeling and inference~\cite{chen2013scalable,paisley2012discrete,putthividhya2009independent,ahmed2007tight}, the model scale has still limited to less than 1000s of latent topics. This stands in stark contrast to recent industrial-scale LDA models which handle millions of topics on billions of documents~\cite{chen2016warplda,yuan2015lightlda} for capturing long-tail semantics and supporting industrial applications~\cite{wang2015peacock},  yet, such rich extraction task is expected to be better addressed with more expressive correlation models. It is therefore highly desirable to develop efficient correlated topic models with great representational power and highly scalable inference, for practical deployment.

\begin{figure*}[!t]
\begin{center}
\includegraphics[scale=0.3]{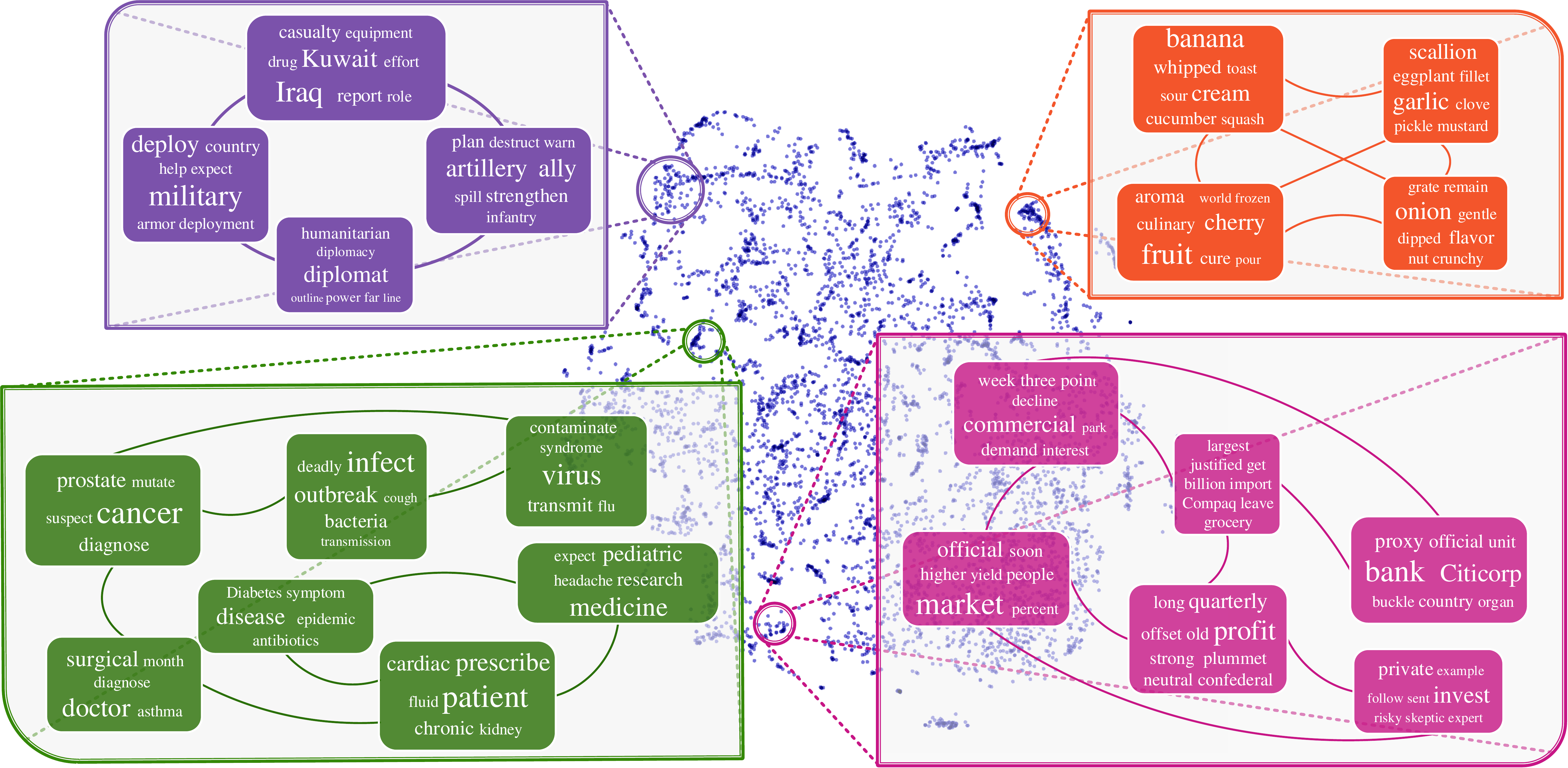}
\vspace{-2pt}
\caption{Visualization of 10K correlated topics on the NYTimes news corpus. The point cloud shows the 10K topic embeddings where each point represents a latent topic. Smaller distance indicates stronger correlation. We show four sets of topics which are nearby each other in the embedding space, respectively. Each topic is characterized by the top words according to the word distribution. Edge indicates correlation between topics with strength above some threshold.}
\label{fig:viz-nytimes}
\end{center}
\vspace{-4pt}
\end{figure*}

In this paper, we develop a new model that extracts correlation structures of latent topics, sharing comparable expressiveness with the costly CTM model, while keeping as efficient as the simple LDA.
We propose to learn a distributed representation for each latent topic, and characterize correlatedness of two topics through the closeness of respective topic vectors in the embedding space. Compared to previous pairwise correlation modeling, our topic embedding scheme is parsimonious with less parameters to estimate, yet flexible to enable richer analysis and visualization. Figure~\ref{fig:viz-nytimes} illustrates the correlation patterns of 10K topics inferred by our model from two million NYTimes news articles, in which we can see clear dependency structures among the large collection of topics and grasp the semantics of the massive text corpus.

We further derive an efficient variational inference procedure combined with a fast sparsity-aware sampler for stochastic tackling of non-conjugacies. Our embedding based correlation modeling enables inference in the low-dimensional vector space, resulting in {\it linear} complexity w.r.t topic size as with the lightweight LDA. This allows us to discover 100s of 1000s of latent topics with their correlations on near 10 million articles, which is several orders of magnitude larger than prior work~\cite{chen2013scalable,blei2007correlated}.

Our work differs from recent research which combines topic models with word embeddings~\cite{li2016generative,batmanghelich2016nonparametric,das2015gaussian,jiang2016latent} for capturing word dependencies, as we instead focus on modeling dependencies in the latent topic space which exhibit uncertainty and are inferentially more challenging. To the best of our knowledge, this is the first work to incorporate distributed representation learning with topic correlation modeling, offering both intuitive geometric interpretation and theoretical Bayesian modeling advantages.  

We demonstrate the efficacy of our method through extensive experiments on various large text corpora. Our approach shows greatly improved efficiency over previous correlated topic models, and scales well as with the much simpler LDA. This is achieved without sacrificing the modeling power---the proposed model extracts high-quality topics and correlations, obtaining competitive or better performance than CTM in document classification and retrieval tasks.  

The rest of the paper is organized as follows: section~\ref{sec:related} briefly reviews related work; section~\ref{sec:model} presents the proposed topic embedding model; section~\ref{sec:exp} shows extensive experimental ; and section~\ref{sec:conclude} concludes the paper.

\section{Related Work}\label{sec:related}
\subsection{Correlated Topic Modeling}
Topic models represent a document as a mixture of latent topics. Among the most popular topic models is the LDA model~\cite{blei2003latent} which assumes conjugate Dirichlet prior over topic mixing proportions for easier inference. Due to its simplicity and scalability, LDA has extracted broad interest for industrial applications~\cite{yuan2015lightlda,wang2015peacock}. 
The Dirichlet prior is however incapable of capturing dependencies between topics. The classic CTM model provides an elegant extension of LDA by replacing the Dirichlet prior with a logistic-normal prior which models pairwise topic correlations with the Gaussian covariance matrix. However, the enriched extraction comes with computational cost. The number of parameters in the covariance matrix grows as square of the number of topics, and parameter estimation for the full-rank matrix can be inaccurate in high-dimensional space. More importantly, frequent matrix inversion operations during inference lead to $\mathcal{O}(K^3)$ time complexity, which has significantly restricted the model and data scales. To address this, \citet{chen2013scalable} derives a scalable Gibbs sampling algorithm based on data augmentation. Though bringing down the inference cost to $\mathcal{O}(K^2)$ per document, the computation is still too expensive to be practical in real-world massive tasks. \citet{putthividhya2009independent} reformulates the correlation prior with independent factor models for faster inference. However, similar to many other approaches, the problem scale has still limited to thousands of documents and hundreds of topics. In contrast, we aim to scale correlated topic modeling to industrial level deployment by reducing the complexity to the LDA level which is linear to the topic size, while providing as rich extraction as the costly CTM model. We note that recent scalable extensions of LDA such as alias methods~\cite{li2014reducing,yuan2015lightlda} are orthogonal to our approach and can be applied in our inference for further speedup. We consider this as our future work. 

Another line of topic models organizes latent topics in a hierarchy which also captures topic dependencies. However, the hierarchy structure is either pre-defined~\cite{li2006pachinko,hu2016grounding,boyd2007topic} or inferred from data using Bayesian nonparametric methods~\cite{dubey2014dependent,blei2010nested} which are known to be computationally demanding~\cite{gal2014pitfalls,hu2015large}. Our proposed model is flexible without sacrificing scalability. 
\subsection{Distributed Representation Learning}
There has been a growing interest in distributed representation that learns compact vectors (a.k.a embeddings) for words~\cite{mikolov2013distributed,lei2014low}, entities~\cite{hu2015entity,li2016joint} , network nodes~\cite{grover2016node2vec,tang2015pte}, and others. The induced vectors are expected to capture semantic relatedness of the target items, and are successfully used in various applications. Compared to most work that induces embeddings for observed units, we learn distributed representations of latent topics which poses unique challenge for inference. Some previous work~\cite{le2014semantic,le2014manifold} also induces compact topic manifold for visualizing large document collections. Our work is distinct in that we leverage the learned topic vectors for efficient correlation modeling and account for the uncertainty of correlations.

An emerging line of approaches~\cite{li2016generative,batmanghelich2016nonparametric,das2015gaussian,jiang2016latent} incorporates word embeddings (either pre-trained or jointly inferred) with conventional topic models for capturing word dependencies and improving topic coherence. Our work differs since we are interested in the topic level, aiming at capturing topic dependencies with learned topic embeddings.

\section{Topic Embedding Model}\label{sec:model}
\begin{figure}
\begin{center}
\includegraphics[scale = 0.35]{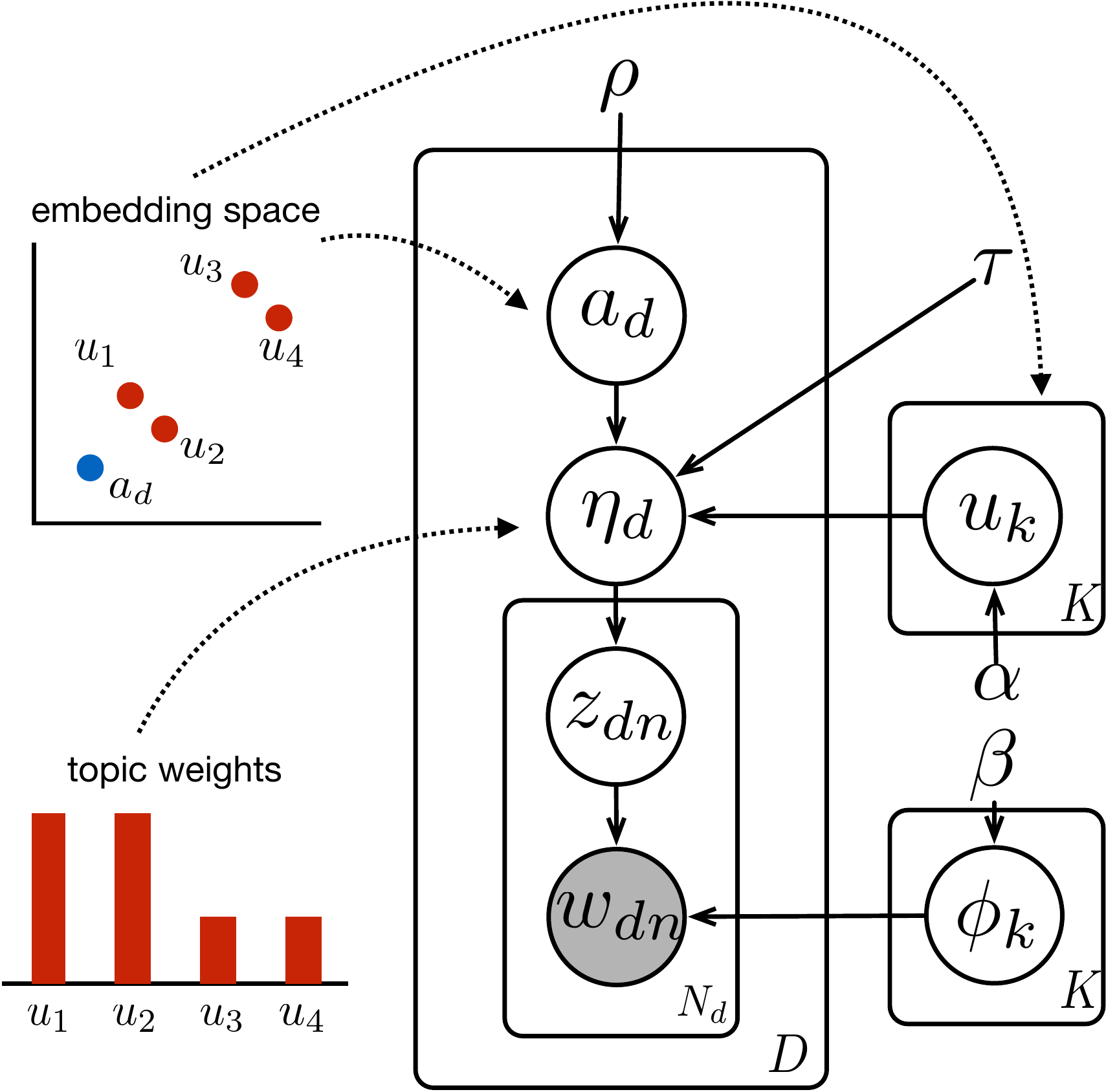}
\caption{Graphical model representation. The left part schematically shows our correlation modeling mechanism, where nearby topics tend to have similar (either large or small) weights in a document.}
\label{fig:graphical}
\end{center}
\vspace{-8pt}
\end{figure}

This section proposes our topic embedding model for correlated topic modeling. We first give an overview of our approach, and present the model structure in detail. We then derive an efficient variational algorithm for inference.  

\subsection{Overview}
We aim to develop an expressive topic model that discovers latent topics and underlying correlation structures. Despite this added representational power, we want to keep the model parsimonious and efficient in order to scale to large text data. As discussed above (section~\ref{sec:related}), CTM captures correlations between topic pairs with a Gaussian covariance matrix, imposing $\mathcal{O}(K^2)$ parameter size and $\mathcal{O}(K^3)$ inference cost. In contrast, we adopt a new modeling scheme drawing inspiration from recent work on distributed representations, such as word embeddings~\cite{mikolov2013distributed} which learn low-dimensional word vectors and have shown to be effective in encoding word semantic relatedness.

We induce continuous distributed representations for latent topics, and, as in word embeddings, expect topics with relevant semantics to be close to each other in the embedding space. The contiguity of the embedding space enables us to capture topical co-occurrence patterns conveniently---we further embed documents into the same vector space, and characterize document's topic proportions with its distances to the topics. Smaller distance indicates larger topic weight. By the triangle inequality of distance metric, intuitively, a document vector will have similar (either large or small) distances to the vectors of two semantically correlated topics which are themselves nearby each other in the space, and thus tend to assign similar probability mass to the two topics. Figure~\ref{fig:graphical}, left part, schematically illustrates the embedding based correlation modeling.

We thus avoid expensive modeling of pairwise topic correlation matrix, and are enabled to perform inference in the low-dimensional embedding space, leading to significant reduction in model and inference complexity. We further exploit the intrinsic sparsity of topic occurrence, and develop stochastic variational inference with fast sparsity-aware sampling to enable high scalability. We derive the inference algorithm in section~\ref{sec:infer}. 

In contrast to word representation learning where word tokens are observed and embeddings can be induced directly from word collocation patterns, topics are hidden from the text, posing additional inferential challenge. 
We resort to generative framework as in conventional topic models by associating a word distribution with each topic. We also take into account uncertainty of topic correlations for flexibility. Thus, in addition to the intuitive geometric interpretation of our embedding based correlation scheme, the full Bayesian treatment also endows connection to the classic CTM model, offering theoretical insights into our approach. We present the model structure in the next section. (Table~\ref{tab:notation} lists key notations; Figure~\ref{fig:graphical} shows the graphical model representation of our model.)

\begin{table}
\centering
\small
\begin{tabular}{@{}r  l@{}}
  \midrule
  Symbol     & Description \\ \cmidrule{1-2} 
  $D, K, V$ & number of documents, latent topics, and vocabulary words \\
  $N_d$ & number of words in document $d$ \\ 
  $M$ & embedding dimension of topic and document \\
  $\bm{u}_k$ & embedding vector of topic $k$ \\
  $\bm{a}_d$ & embedding vector of document $d$ \\
  $\bm{\eta}_d$ & (unnormalized) topic weight vector of document $d$ \\
  $w_{dn}$ & the $n$th word in document $d$ \\
  $z_{dn}$ & the topic assignment of word $w_{dn}$ \\
  $\bm{\phi}_k$ & word distribution of topic $k$ \\
  $K_s$ & number of non-zero entries of document's topic proportion \\
  $V_s$ & number of non-zero entries of topic word distribution \\ 
  \midrule
\end{tabular}
\caption{Notations used in this paper.}
\label{tab:notation}
\vspace{-20pt}
\end{table}

\subsection{Model Structure}\label{sec:struct}

We first establish the notations. Let $\bm{W} = \{\bm{w}_d\}_{d=1}^{D}$ be a collection of documents. Each document $d$ contains $N_d$ words $\bm{w}_d = \{w_{dn}\}_{n=1}^{N_d}$ from a vocabulary of size $V$. 

We assume $K$ topics underlying the corpus. As discussed above, for each topic $k$, we want to learn a compact distributed representation $\bm{u}_k \in \mathbb{R}^{M}$ with low dimensionality ($M\ll K$). Let $\bm{U}\in \mathbb{R}^{K\times M}$ denote the topic vector collection with the $k$th row $\bm{U}_{k\cdot} = \bm{u}_k^{T}$. As a common choice in word embedding methods, we use the vector inner product for measuring the closeness between embedding vectors. 
In addition to topic embeddings, we also induce document vectors in the same vector space. Let $\bm{a}_d \in \mathbb{R}^{M}$ denote the embedding of document $d$. We now can conveniently compute the affinity of a document $d$ to a topic $k$ through $\bm{u}_k^T\bm{a}_d$. A topic $k'$ nearby, and thus semantically correlated to topic $k$, will naturally have similar distance to the document, as $|\bm{u}_k^T\bm{a}_d - \bm{u}_{k'}^T\bm{a}_d| \leq \|\bm{u}_k-\bm{u}_{k'}\|\|\bm{a}_d\|$ and $\|\bm{u}_k-\bm{u}_{k'}\|$ is small.

We express uncertainty of the affinity by modeling the actual topic weights $\bm{\eta}_d \in \mathcal{R}^{K}$ as a Gaussian variable centered at the affinity vector, following $\bm{\eta}_d \sim \mathcal{N}(\bm{U}\bm{a}_d, \tau^{-1}\bm{I})$. Here $\tau$ characterizes the uncertainty degree and is pre-specified for simplicity. 
As in logistic-normal models, we project the topic weights into the probability simplex to obtain topic distribution $\bm{\theta}_{d} = \text{softmax}(\bm{\eta}_d)$, from which we sample a topic $z_{dn}\in\{1,\dots,K\}$ for each word $w_{dn}$ in the document. As in conventional topic models, each topic $k$ is associated with a multinomial distribution $\bm{\phi}_k$ over the word vocabulary, and each observed word is drawn from respective word distribution indicated by its topic assignment.

Putting everything together, the generative process of the proposed model is summarized in Algorithm~\ref{algo:gen}. A theoretically appealing property of our method is its intrinsic connection to conventional logistic-normal models such as the CTM model. If we marginalize out the document embedding variable $\bm{a}_d$, we obtain $\bm{\eta}_d \sim \mathcal{N}(\bm{0}, \bm{U}\bm{U}^{T}+\tau^{-1}\bm{I})$, recovering the pairwise topic correlation matrix with low rank constraint, where each element is just the closeness of respective topic embeddings, coherent to the above geometric intuitions. Such covariance decomposition has been used in other context, such as sparse Gaussian processes~\cite{titsias2009variational} for efficient approximation and Gaussian reparameterization~\cite{kingma2013auto,wilson2016stochastic} for differentiation and reduced variance. Here we relate low-dimensional embedding learning with low-rank covariance decomposition and estimation.

The low-dimensional representations of latent topics enable parsimonious correlation modeling with parameter complexity of $\mathcal{O}(MK)$ (i.e., topic embedding parameters), which is efficient in terms of topic number $K$. Moreover, we are allowed to perform efficient inference in the embedding space, with inference cost {\it linear} in $K$, a huge advance compared to previous cubic complexity of vanilla CTM~\cite{blei2007correlated} and quadratic of recent improved version~\cite{chen2013scalable}. We derive our inference algorithm in the next section.

\subsection{Inference}\label{sec:infer} 
Posterior inference and parameter estimation is not analytically tractable due to the coupling between latent variables and the non-conjugate logistic-normal prior. This makes the learning difficult especially in our context of scaling to unprecedentedly large data and model sizes.
We develop a stochastic variational method that (1) involves only compact topic vectors which are cheap to infer, and (2) includes a fast sampling strategy which tackles non-conjugacy and exploits intrinsic sparsity of both the document topic occurrence and the topical words.

We first assume a mean-field family of variational distributions:
\begin{equation}
\begin{split}
&q(\bm{u},\bm{\phi},\bm{a},\bm{\eta},\bm{z}) = \\ &\prod\nolimits_{k}q(\bm{u}_k)q(\bm{\phi}_k) \prod\nolimits_{d}q(\bm{a}_d)q(\bm{\eta}_d)\prod\nolimits_n q(z_{dn}).
\end{split}
\end{equation}
where the factors have the parametric forms:
 \begin{equation}
 \begin{split}
 q(\bm{u}_k)=\mathcal{N}(\bm{u}_k|\bm{\mu}_k, \Sigma^{(u)}_{k}),&\quad q(\bm{a}_d)=\mathcal{N}(\bm{a}_d|\bm{\gamma}_d, \Sigma^{(a)}_{d}),\\
 q(\bm{\phi}_k) = \text{Dir}(\bm{\phi}_k|\bm{\lambda}_k),&\quad q(\bm{\eta}_d) = \mathcal{N}(\bm{\eta}_d|\bm{\xi}_d, \Sigma^{(\eta)}_{d}), \\
 q(z_{dn}) = \text{Multi}(z_{dn}|\bm{\kappa}_{dn})&
\end{split}
\label{eq:vi-factor}
\end{equation}
Variational algorithms aim to minimize KL divergence from $q$ to the true posterior, which is equivalent to tightening the evidence lower bound (ELBO):
\begin{equation}
\begin{split}
\mathcal{L}(q) = &\sum\nolimits_k \E_{q}\left[ \log \frac{p(\bm{u}_k)p(\bm{\phi}_k)}{q(\bm{u}_k)q(\bm{\phi}_k)}  \right] + \\
&\sum\nolimits_{d,n} \E_{q}\left[ \log \frac{p(\bm{a}_d) p(\bm{\eta}_d | \bm{a}_d, \bm{U}) p(z_{dn}|\bm{\eta}_d) p(w_{dn}|z_{dn},\bm{\phi})}{q(\bm{a}_d)q(\bm{\eta}_d)q(z_{dn})} \right] 
\end{split}
\end{equation}
We optimize $\mathcal{L}(q)$ via coordinate ascent, interleaving the update of the variational parameters at each iteration. We employ stochastic variational inference which optimizes the parameters with stochastic gradients estimated on data minibatchs. 
Due to the space limitations, here we only describe key computation rules of the gradients (or closed-form solutions). These stochastically estimated quantities are then used to update the variational parameters after scaled by a learning rate. Please refer to the supplementary material~\cite{supplement} for detailed derivations.

\begin{algorithm}[t]
\begin{enumerate}[1.]
   \item For each topic $k = 1, 2, \cdots, K$,
  \begin{itemize}
   \item Draw the topic word distribution $\bm{\phi_k} \sim \text{Dir}(\beta)$ 
   \item Draw the topic embedding $\bm{u}_k \sim \mathcal{N}(\bm{0}, \alpha^{-1}\bm{I})$
  \end{itemize}
  \item For each document $d = 1, 2, \cdots, D$,
   \begin{itemize}
    \item Draw the document embedding $\bm{a}_d \sim \mathcal{N}(\bm{0}, \rho^{-1}\bm{I})$
    \item Draw the document topic weight $\bm{\eta}_d \sim \mathcal{N}(\bm{U}\bm{a}_d, \tau^{-1}\bm{I})$
    \item Derive the distribution over topics $\bm{\theta}_d = \text{softmax}(\bm{\eta}_d)$
    \item For each word $n = 1, 2, \cdots, N_d$,\par
    (a) Draw the topic assignment $z_{dn} \sim \text{Multi}(\bm{\theta}_d)$\par
    (b) Draw the word $w_{dn} \sim \text{Multi}(\bm{\phi}_{z_{dn}})$
   \end{itemize}
\end{enumerate}
\caption{Generative Process}
\label{algo:gen}
\end{algorithm}

{\bf Updating topic and document embeddings. }
For each topic $k$, we isolate only the terms that contain $q(\bm{u}_k | \bm{\mu}_k, \Sigma^{(u)}_k)$,
\begin{equation}
\begin{split}
\mathcal{L}(q(\bm{u}_k)) =\  &\E_{q}\left[\log p(\bm{u}_k)\right] + \sum\nolimits_{d}\E_{q}\left[\log p(\bm{\eta}_d | \bm{a}_d, \bm{U})\right] \\
&- \E_q \left[ \log q(\bm{u}_k) \right].
\end{split}
\end{equation}
The optimal solution for $q(\bm{u}_k)$ is then obtained by setting the gradient to zero, with the variational parameters computed as: 
\begin{equation}
\begin{split}
\bm{\mu}_k &= \tau\Sigma^{(u)}\cdot\left( \sum\nolimits_{d} \bm{\xi}_{dk}\bm{\gamma}_d \right), \\
\Sigma^{(u)} &= \left[\alpha\bm{I} + \tau \sum\nolimits_{d}\left(\Sigma^{(a)}_d + \bm{\gamma}_d\bm{\gamma}_d^T\right)\right]^{-1},
\end{split}
\label{eq:vi-u}
\end{equation}
where we have omitted the subscript $k$ of the variational covariance matrix $\Sigma^{(u)}$ as it is independent with $k$. Intuitively, the optimal variational topic embeddings are the centers of variational document embeddings scaled by respective document topic weights and transformed by the variational covariance matrix.  

By symmetry, the variational parameters of document embedding $\bm{a}_d$ is similarly updated as:
\begin{equation}
\begin{split}
\bm{\gamma}_d &= \tau\Sigma^{(a)}\cdot\left( \sum\nolimits_{k} \xi_{dk}\bm{\mu}_k \right), \\
\Sigma^{(a)} &= \left[\gamma\bm{I} + \tau \sum\nolimits_{k}\left(\Sigma^{(u)} + \bm{\mu}_k\bm{\mu}_k^T\right)\right]^{-1},
\end{split}
\label{eq:vi-a}
\end{equation}
where, again, $\Sigma^{(a)}$ is independent with $d$ and thus the subscript $d$ is omitted.

Learning low-dimensional topic and document embeddings is computationally cheap. Specifically, by Eq.\eqref{eq:vi-u}, updating the set of variational topic vector means $\{\bm{\mu}_k\}_{k=1}^{K}$ imposes complexity $\mathcal{O}(KM^2)$, and updating the covariance $\Sigma^{(u)}$ requires only $\mathcal{O}(M^3)$. Similarly, by Eq.\eqref{eq:vi-a}, the cost of optimizing $\bm{\gamma}_d$ and $\Sigma^{(a)}$ is $\mathcal{O}(KM)$ and $\mathcal{O}(KM^2)$, respectively. Note that $\Sigma^{(a)}$ is shared across all documents and does not need updates per document. We see that all the updates cost only linearly w.r.t to the topic size $K$ which is critical to scale to large-scale practical applications.

{\bf Sparsity-aware topic sampling. }
We next consider the optimization of the variational topic assignment $q(z_{dn})$ for each word $w_{dn}$. Letting $w_{dn}=v$, the optimal solution is:
\begin{equation}
\begin{split}
q(z_{dn}=k) \propto \exp\left\{ \bm{\xi}_{dk} \right\}
\exp\left\{ \Psi(\lambda_{kv}) - \Psi\left(\sum\nolimits_{v'}\lambda_{kv'}\right) \right\},
\end{split}
\label{eq:vi-zdn}
\end{equation}
where $\Psi(\cdot)$ is the digamma function; and $\bm{\xi}_d$ and $\bm{\lambda}_k$ are the variational means of the document's topic weights and the variational word weights (Eq.\eqref{eq:vi-factor}), respectively. Direct computation of $q(z_{dn})$ with Eq.\eqref{eq:vi-zdn} has complexity of $\mathcal{O}(K)$, which becomes prohibitive in the presence of many latent topics. To address this, we exploit two aspects of intrinsic sparsity in the modeling: (1) Though a whole corpus can cover a large diverse set of topics, a single document in the corpus is usually about only a small number of them. We thus only maintain the top $K_s$ entries in each $\bm{\xi}_d$, where $K_s \ll K$, 
making the complexity due to the first term in the right-hand side of Eq.\eqref{eq:vi-zdn} only $\mathcal{O}(K_s)$ for all $K$ topics in total; 
(2) A topic is typically characterized by only a few words in the large vocabulary, we thus cut off the variational word weight vector $\bm{\lambda}_k$ for each $k$ by maintaining only its top $V_s$ entries ($V_s \ll V$). Such sparse treatment helps enhance the interpretability of learned topics, 
and allows cheap computation with on average $\mathcal{O}(KV_s/V)$ cost for the second term\footnote{In practice we also set a threshold $s$ such that each word $v$ needs to have at least $s$ non-zero entries in $\{\lambda_{k}\}_{k=1}^{K}$. Thus the exact complexity of the second term is $\mathcal{O}(\max\{KV_s/V, s\})$.}.
With the above sparsity-aware updates, the resulting complexity for Eq.\eqref{eq:vi-zdn} with $K$ topics is brought down to  $\mathcal{O}(K_s+KV_s/V)$, a great speedup over the original $\mathcal{O}(K)$ cost. The top $K_s$ entries of $\bm{\xi}_d$ are selected using a Min-heap data structure, whose computational cost is amortized across all words in the document, imposing $\mathcal{O}(K/N_d\log K_s)$ computation per word. The cost for finding the top $V_s$ entries of $\bm{\lambda}_k$ is similarly amortized across documents and words, and becomes insignificant.

Updating the remaining variational parameters will frequently involve computation of variational expectations under $q(z_{dn})$. It is thus crucial to speedup this operation. To this end, we employ sparse approximation by sampling from $q(z_{dn})$ a single indicator $\tilde{z}_{dn}$, and use the ``hard'' sparse distribution $\tilde{q}(z_{dn}=k):=\bm{1}(\tilde{z}_{dn}=k)$ to estimate the expectations. Note that the sampling operation is cheap, having the same complexity with computing $q(z_{dn})$ as above. As shown shortly, such sparse computation will significantly reduce our running cost. Though stochastic expectation approximation is commonly used for tackling intractability~\cite{mimno2012sparse,lazaro2014doubly}, here we instead apply the technique for fast estimation of tractable expectations.

We next optimize the variational topic weights $q(\bm{\eta}_d | \bm{\xi}_d, \Sigma^{(\eta)}_d)$. Extracting only the terms in $\mathcal{L}(q)$ involving $q(\bm{\eta}_d)$, we get:
\begin{equation}\label{eq:elbo-q-eta}
\begin{split}
\mathcal{L}(q(\bm{\eta}_d)) =\  &\E_q\left[\log p(\bm{\eta}_d | \bm{a}_d, \bm{U}) \right] + \E_q\left[ \log p(\bm{z}_{d} | \bm{\eta}_d) \right] \\
&- \E_q\left[\log q(\bm{\eta}_d) \right],
\end{split}
\end{equation}
where the second term 
\begin{equation*}
\begin{split}
\E_q\left[ \log p(\bm{z}_{d} | \bm{\eta}_d) \right] = \sum\nolimits_{k,n} q(z_{dn} = k) \E_q\left[\log (\text{softmax}_k(\bm{\eta}_d))\right] 
\end{split}
\end{equation*}
involves variational expectations of the logistic transformation which does not have an analytic form. We construct a fast Monto Carlo estimator for approximation. Particularly, we employ reparameterization trick by first assuming a diagonal covariance matrix $\Sigma^{(\eta)}_d = \text{diag}(\bm{\sigma}_d^2)$ as is commonly used in previous work~\cite{blei2007correlated,kingma2013auto}, where $\bm{\sigma}_d$ denotes the vector of standard deviations, resulting in the following sampling procedure:
\begin{equation}
\begin{split}
\bm{\eta}^{(t)}_d = \bm{\xi}_d + \bm{\sigma}_d \odot \bm{\epsilon}^{(t)};\quad \bm{\epsilon}^{(t)} \sim \mathcal{N}(\bm{0},\bm{I}),
\end{split}
\end{equation}
where $\odot$ is the element-wise multiplication.
With $T$ samples of $\bm{\eta}_d$, we can estimate the variational lower bound and the derivatives $\nabla\mathcal{L}$ w.r.t the variational parameters $\{\bm{\xi}_d, \bm{\sigma}_d\}$. For instance, 
\begin{equation}
\begin{split}
&\nabla_{\bm{\xi}_d} \mathbb{E}_q \left[\log p(\bm{z}_{d} | \bm{\eta}_d) \right] \\
&\approx \sum\nolimits_{k,n}q(z_{dn}=k) \bm{e}_k - (N_d/T)\sum\nolimits_{t=1}^{T}\textbf{softmax}\left(\bm{\eta}_{d}^{(t)}\right) \\
&\approx \sum\nolimits_{k,n}\bm{1}(\tilde{z}_{dn}=k) \bm{e}_k - (N_d/T)\sum\nolimits_{t=1}^{T}\textbf{softmax}\left(\bm{\eta}_{d}^{(t)}\right)
\end{split}
\label{eq:vi-p-zdn}
\end{equation}
where $\bm{e}_k$ is an indicator vector with the $k$th element being $1$ and the rest $0$. In practice $T=1$ is usually sufficient for effective inference. The second equation applies the hard topic sample mentioned above, which reduces the time complexity $\mathcal{O}(KN_d)$ of the original standard computation (the first equation) to $\mathcal{O}(N_d+K)$ (i.e., $\mathcal{O}(N_d)$ for the first term and $\mathcal{O}(K)$ for the second).

The first term in Eq.\eqref{eq:elbo-q-eta} depends on the topic and document embeddings to encode topic correlations in document's topic weights. The derivative w.r.t to the variational parameter $\bm{\xi}_d$ is computed as:
\begin{equation}
\nabla_{\bm{\xi}_d}\mathbb{E}_q\left[\log p(\bm{\eta}_d | \bm{U}, \bm{a}_d)\right] = \tau(\tilde{\bm{U}} \bm{\gamma}_d - \bm{\xi}_d).
\label{eq:vi-p-eta}
\end{equation}
Here $\tilde{\bm{U}}$ is the collection of variational means of topic embeddings where the $k$th row $\tilde{\bm{U}}_{k\cdot} = \bm{\mu}_k^T$. We see that, with low-dimensional topic and document vector representations, inferring topic correlations is of low cost $\mathcal{O}(KM)$ which grows only linearly w.r.t to the topic size.
The complexity of the remaining terms in Eq.\eqref{eq:elbo-q-eta}, as well as respective derivatives w.r.t the variational parameters, has complexity of $\mathcal{O}(KM)$ (Please see the supplements~\cite{supplement} for more details). In summary, the cost of updating $q(\bm{\eta}_d)$ for each document $d$ is $\mathcal{O}(KM+K+N_d)$. 

Finally, the optimal solution of the variational topic word distribution $q(\bm{\phi}_k | \bm{\lambda}_k)$ is given by:
\begin{equation}
\begin{split}
\lambda_{kv} = \beta + \sum\nolimits_{d,n}\bm{1}(w_{dn}=v) \bm{1}(\tilde{z}_{dn}=k).
\end{split}
\label{eq:vi-lambda}
\end{equation}

\begin{algorithm}[t]
\centering
\caption{\small Stochastic variational inference}
\label{alg:opt}
\begin{algorithmic}[1]
\STATE Initialize variational parameters randomly
\REPEAT
    \STATE Compute learning rate $\iota_{\textit{iter}}=1/(1+\textit{iter})^{0.9}$
    \STATE Sample a minibatch of documents $\mathcal{B}$
    \FORALL {$d\in \mathcal{B}$}
    \REPEAT
    	\STATE Update $q(\bm{z}_d)$ with Eq.\eqref{eq:vi-zdn} and sample $\tilde{\bm{z}}_d$
        \STATE Update $\bm{\gamma}_d$ with Eq.\eqref{eq:vi-a}
        \STATE Update $q(\bm{\eta}_d)$ using respective gradients computed with Eqs.\eqref{eq:vi-p-zdn},\eqref{eq:vi-p-eta},and more in the supplements~\cite{supplement}.
    \UNTIL{convergence}
    \STATE Compute stochastic optimal values $\bm{\mu}^{*}, \Sigma^{(u)*}$ with Eq.\eqref{eq:vi-u}
    \STATE Compute stochastic optimal values $\bm{\lambda}^{*}$ with Eq.\eqref{eq:vi-lambda}
    \STATE Update $\bm{x} = (1-\iota_\textit{iter})\bm{x}+\iota_\textit{iter}\bm{x}^{*}$ with $\bm{x}\in\{\bm{\mu}, \Sigma^{(u)}, \bm{\lambda}\}$
    \STATE Update $\Sigma^{(a)}$ with Eq.\eqref{eq:vi-a}
    \ENDFOR
\UNTIL{convergence}
\end{algorithmic}
\end{algorithm}

{\bf Algorithm summarization. } 
We summarize our variational inference in Algorithm~\ref{alg:opt}. 
As analyzed above, the time complexity of our variational method is $\mathcal{O}(KM^2+M^3)$ for inferring topic embeddings $q(\bm{u}_{d})$. The cost per document is $\mathcal{O}(KM)$ for computing $q(\bm{a}_d)$, $\mathcal{O}(KM)$ for updating $q(\bm{\eta}_d)$, and $\mathcal{O}((K_s+KV_s/V)N_d)$ for maintaining $q(\bm{z}_{d})$. The overall complexity for each document is thus $\mathcal{O}(KM+(K_s+KV_s/V)N_d)$, which is linear to model size ($K$), comparable to the LDA model while greatly improving over previous correlation methods with cubic or quadratic complexity. 

The variational inference algorithm endows rich independence structures between the variational parameters, allowing straightforward parallel computing. In our implementation, updates of variational topic embeddings $\{\bm{\mu}_k\}$ (Eq.\eqref{eq:vi-u}), topic word distributions $\{\bm{\lambda}_k\}$ (Eq.\eqref{eq:vi-lambda}), and document embeddings $\{\bm{\gamma}_d\}$ (Eq.\eqref{eq:vi-a}) for a data minibatch, are all computed in parallel across multiple CPU cores.

\section{Experiments}\label{sec:exp}
We demonstrate the efficacy of our approach with extensive experiments. (1) We evaluate the extraction quality in the tasks of document classification and retrieval, in which our model achieves similar or better performance than existing correlated topic models, significantly improving over simple LDA. (2) For scalability, our approach scales comparably with LDA, and handles massive problem sizes orders-of-magnitude larger than previously reported correlation results. (3) Qualitatively, our model reveals very meaningful topic correlation structures.

\subsection{Setup}
\textbf{Datasets. }We use three public corpora provided in the UCI repository\footnote{http://archive.ics.uci.edu/ml} for the evaluation: {\bf 20Newsgroups} is a collection of news documents partitioned (nearly) evenly across 20 different newsgroups. Each article is associated with a category label, serving as ground truth in the tasks of document classification and retrieval; {\bf NYTimes} is a widely-used large corpus of New York Times news articles; and {\bf PubMed} is a large set of PubMed abstracts.
The detailed statistics of the datasets are listed in Table~\ref{tb:data}. We removed a standard list of 174 stop words and performed stemming. For NYTimes and Pubmed, we kept the top 10K frequent words in vocabulary, and selected 10\% documents uniformly at random as test sets, respectively. For 20Newsgroups, we followed the standard training/test splitting, and performed the widely-used pre-processing\footnote{http://scikit-learn.org/stable/datasets/twenty\_newsgroups.html} by removing indicative meta text such as headers and footers so that document classification is forced to be based on the semantics of plain text.

\begin{table}[!t]
\centering
\begin{tabular}{l r r r}
\cmidrule[\heavyrulewidth]{1-4}
Dataset  & \#doc ($D$)  & vocab size ($V$) & doc length \\
\cmidrule{1-4}
20Newsgroups  & 18K & 30K & 130\\
NYTimes     & 1.8M & 10K & 284 \\
PubMed    & 8.2M & 10K & 77\\
\cmidrule[\heavyrulewidth]{1-4}
\end{tabular}
\caption{Statistics of the three datasets, including the number of documents ($D$), vocabulary size ($V$), and average number of words in each document.} 
\label{tb:data}
\vspace{-14pt}
\end{table}
\noindent\textbf{Baselines. }
We compare the proposed model with a set of carefully selected competitors:
\begin{itemize}
\item  {\bf Latent Dirichlet Allocation (LDA)}~\cite{blei2003latent} uses conjugate Dirichlet priors and thus scales linearly w.r.t the topic size but fails to capture topic correlations. Inference is based on the stochastic variational algorithm~\cite{hoffman2013stochastic}. When evaluating scalability, we leverage the same sparsity assumptions as in our model for speeding up.
\item {\bf Correlated Topic Model (CTM)}~\cite{blei2007correlated} employs standard logistic-normal prior which captures pairwise topic correlations. The model uses stochastic variational inference with $\mathcal{O}(K^3)$ time complexity.
\item {\bf Scalable CTM (S-CTM)}~\cite{chen2013scalable} developed a scalable sparse Gibbs sampler for CTM inference with time complexity of $\mathcal{O}(K^2)$. Using distributed inference on 40 machines, the method discovers 1K topics from millions of documents, which to our knowledge is the largest automatically learned topic correlation structures so far.
\end{itemize}

\noindent\textbf{Parameter Setting. }Throughout the experiments, we set the embedding dimension to $M=50$, and sparseness parameters to $K_s = 50$ and $V_s = 100$. We found our modeling quality is robust to these parameters. Following common practice, the hyper-parameters are fixed to $\beta = 1/K, \alpha = 0.1, \rho = 0.1$, and $\tau=1$. The baselines are using similar hyper-parameter settings.

All experiments were performed on Linux with 24 4.0GHz CPU cores and 128GB RAM. All models are implemented using C/C++, and parallelized whenever possible using the OpenMP library. 

\begin{figure}[!t]
\begin{center}
\includegraphics[scale = 0.6]{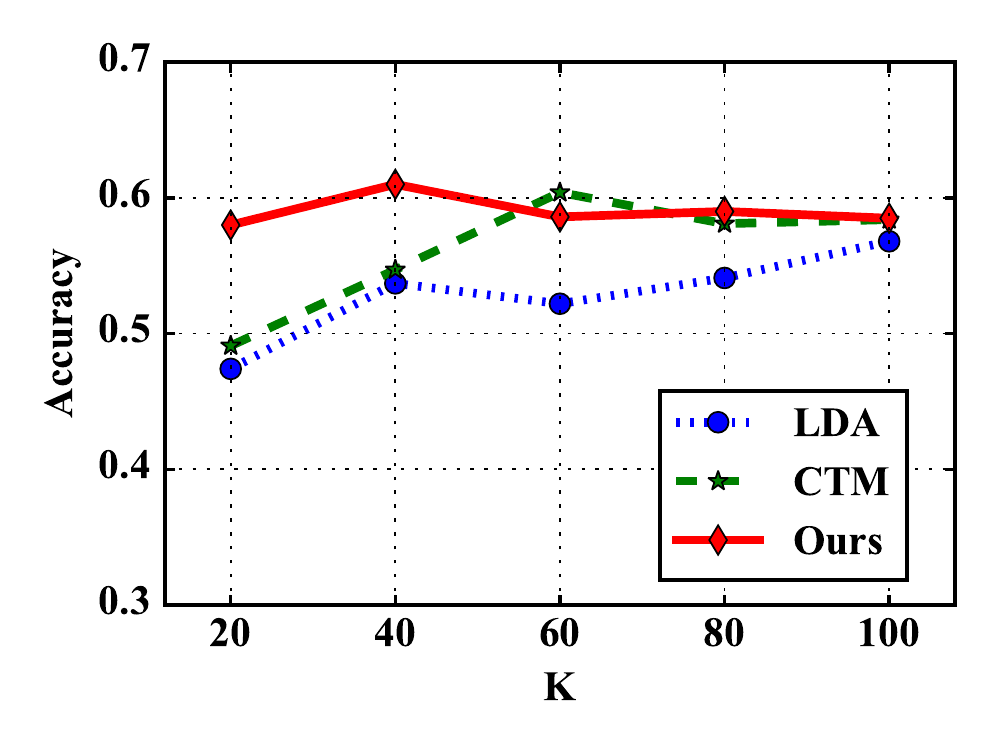}
\vspace{-15pt}
\caption{Classification accuracy on 20newsgroup.}
\label{fig:cls-20news}
\end{center}
\vspace{-10pt}
\end{figure}

\begin{figure*}[!t]
\begin{center}
\subfigure{
\includegraphics[scale=0.52]{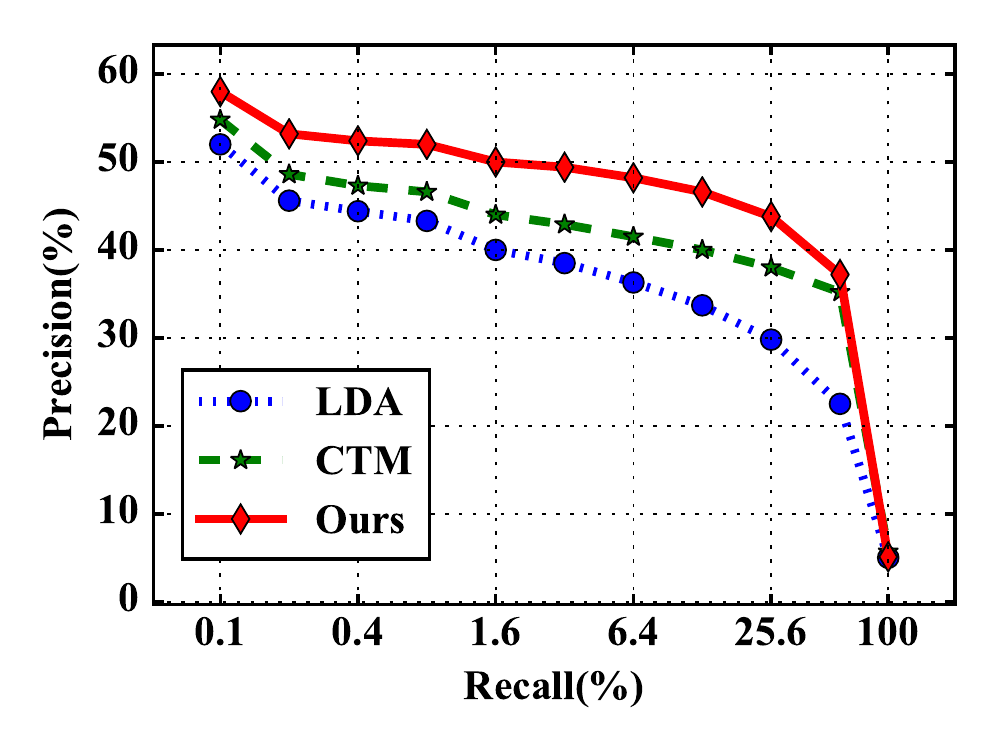}
}
\hfil
\subfigure{
\includegraphics[scale=0.52]{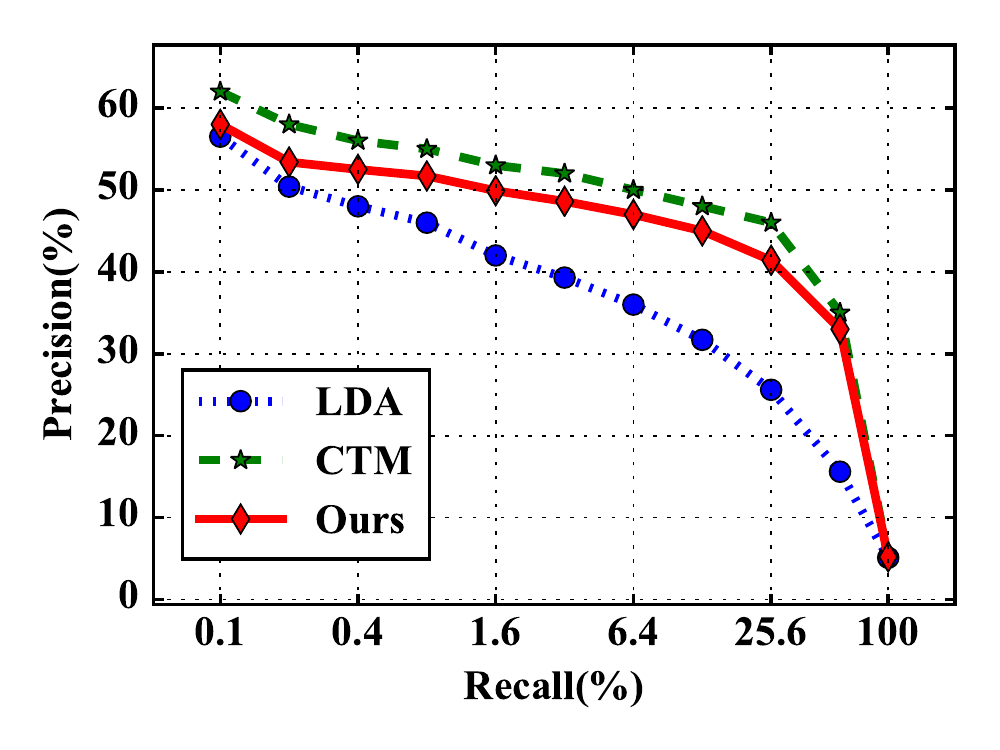}
}
\hfil
\subfigure
{\includegraphics[scale=0.52]{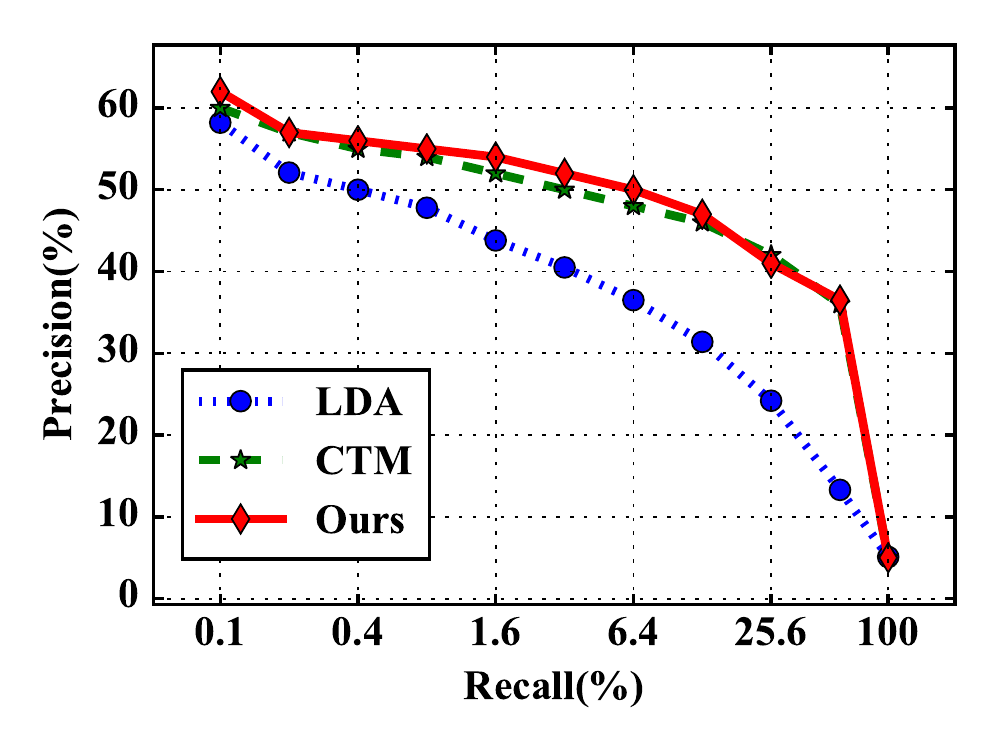}
}
\vspace{-13pt}
\caption{Precision-Recall curves on 20Newsgroups. Left: \#topic $K=20$. Middle: $K=60$. Right: $K=100$.}
\label{fig:doc-ir}
\end{center}
\vspace{-3pt}
\end{figure*}

\subsection{Document Classification} \label{sec:cls}
We first evaluate the performance of document classification based on the learned document representations. We evaluate on the 20Newsgroups dataset where ground truth class labels are available. We compare our proposed model with LDA and CTM. For LDA and CTM, a multi-class SVM classifier is trained for each of them based on the topic distributions of the training documents, while for the proposed model, the SVM classifier takes the document embedding vectors as input. Generally, more accurate modeling of topic correlations enables better document modeling and representations, resulting in improved document classification accuracy.

Figure~\ref{fig:cls-20news} shows the classification accuracy as the number of topics varies. We see that the proposed model performs best in most of the cases, indicating that our method can discover high-quality latent topics and correlations. Both CTM and our model significantly outperforms LDA which treats latent topics independently, validating the importance of topic correlation for accurate text semantic modeling. Compared to CTM, our method achieves better or competitive accuracy as $K$ varies, which indicates that our model, though orders-of-magnitude faster (as shown in the next), does not sacrifice modeling power compared to the complicated and computationally demanding CTM model.

\subsection{Document Retrieval} \label{sec:ir}
We further evaluate the topic modeling quality by measuring the performance of document retrieval~\cite{hinton2009replicated}. We use the 20Newsgroups dataset. A retrieved document is relevant to the query document when they have the same class label. For LDA and CTM, document similarity is measured as the inner product of topic distributions, and for our model we use the inner product of document embedding vectors.  

Figure~\ref{fig:doc-ir} shows the retrieval results with varying number of topics, where we use the test set as query documents to retrieve similar documents from the training set, and the results are averaged over all possible queries. We observe similar patterns as in the document classification task. Our model obtains competitive performance with CTM, both of which capture topic correlations and greatly improve over LDA. This again validates our goal that the proposed method has lower modeling complexity while at the same time is as accurate and powerful as previous complicated correlation models. 
In addition to efficient model inference and learning, our approach based on compact document embedding vectors also enables faster document retrieval compared to conventional topic models which are based on topic distribution vectors (i.e., $M\ll K$). 

\begin{figure*}[!t]
\begin{center}
\subfigure
{\includegraphics[scale=0.54]{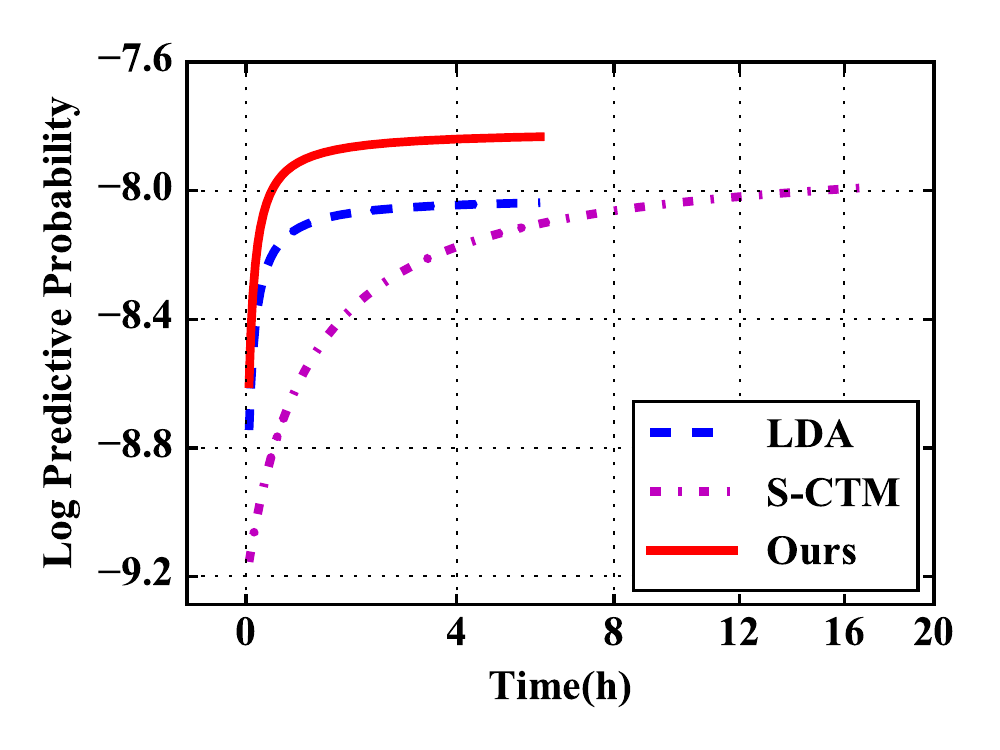}
}
{\includegraphics[scale=0.52]{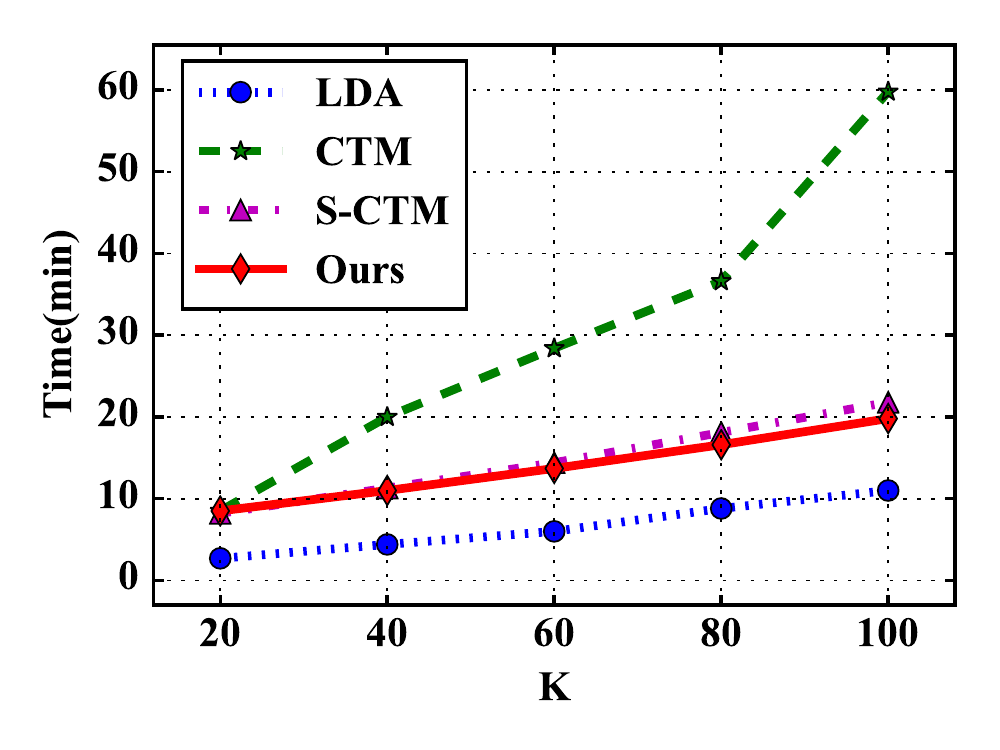}
}
{\includegraphics[scale=0.52]{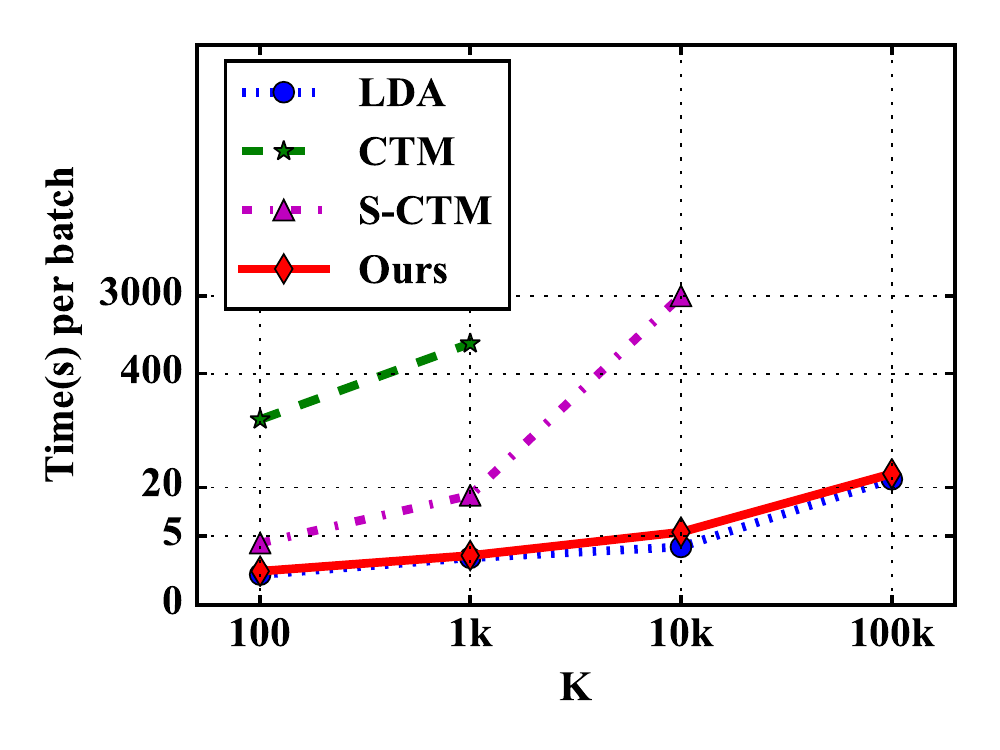}
}
\vspace{-15pt}
\caption{Left: Convergence on NYTimes with 1K topics. Middle: Total training time on 20Newsgroups. Right: Runtime of one  inference iteration on a minibatch of 500 NYTimes articles, where the result points of CTM and S-CTM on large $K$ are omitted as they fail to finish one iteration within $2$ hours.}
\label{fig:scale}
\end{center}
\end{figure*}
\begin{figure*}[!h]
\begin{center}
\includegraphics[scale=0.27]{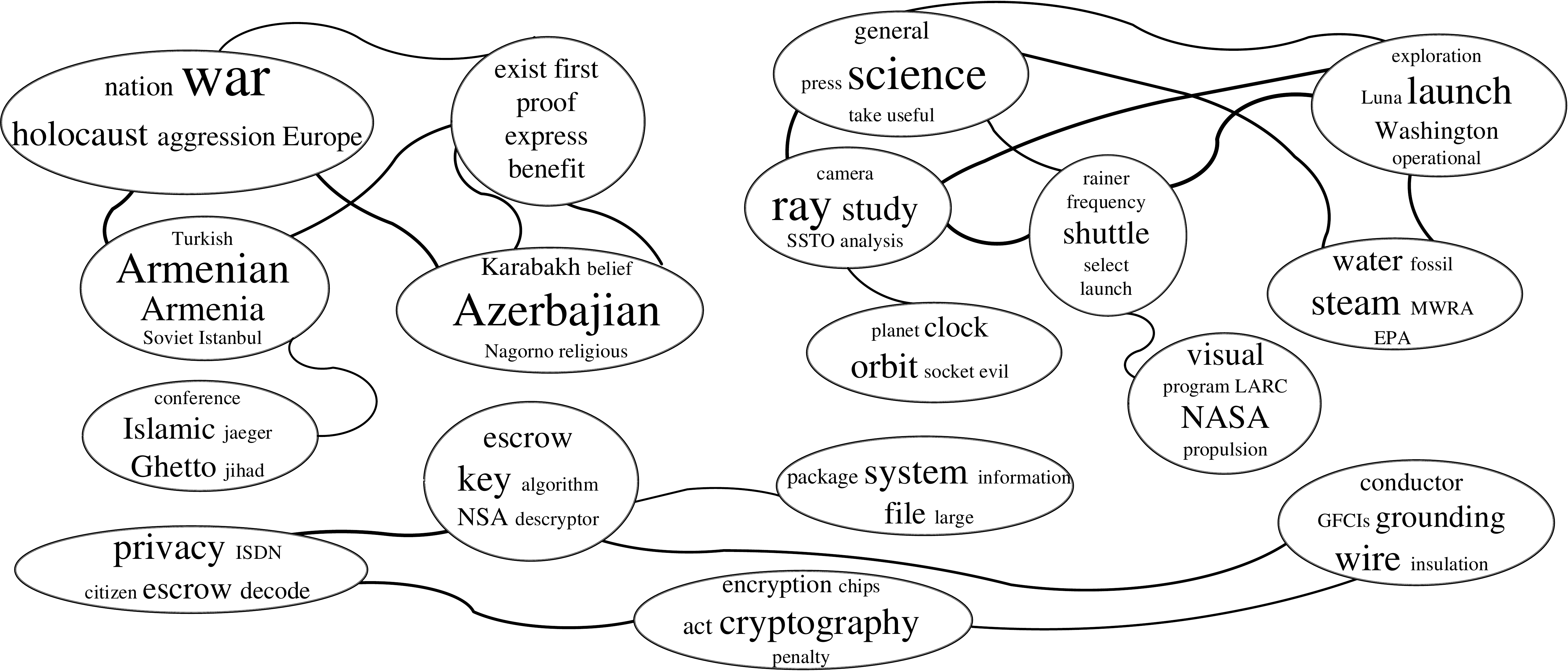}
\caption{A portion of topic correlation graph learned from 20Newsgroups. Each node denotes a latent topic whose semantic meaning is characterized by the top words according to the topic's word distribution. The font size of each word is proportional to the word weight. Topics with correlation strength over some threshold are connected with edges. The thickness of the edges is proportional to the correlation strengths.}
\label{fig:viz-20news}
\end{center}
\end{figure*}
\subsection{Scalability}
We now investigate the efficiency and scalability of the proposed model. Compared to topic extraction quality in which our model achieves similar or better level of performance as the conventional complicated correlated topic model, here we want our approach to tackle large problem sizes which are impossible for existing correlation methods, and to scale as efficiently as the lightweight LDA, for practical deployment.  

\begin{table}[!t]
\centering
\begin{tabular}{l r r r r r}
\cmidrule[\heavyrulewidth]{1-6}
\multirow{2}{*}{Dataset}  & \multirow{2}{*}{$K$}  & \multicolumn{4}{c}{Running Time} \\ \cmidrule(l){3-6}
 & & LDA & CTM & S-CTM & Ours \\ 
\cmidrule[\heavyrulewidth]{1-6}
20Newsgroups  & 100 & 11 min & 60 min & 22 min & 20 min\\
\cmidrule{1-6}
\multirow{3}{*}{NYTimes}     & 100 & 2.5 hr & -- & 6.4 hr & 3.5 hr \\
     & 1K & 5.6 hr & -- & -- & 5.7 hr \\
     & 10K & 8.4 hr & -- & -- & 9.2 hr \\
\cmidrule{1-6}
PubMed     & 100K & 16.7 hr & -- & -- & 19.9 hr \\
\cmidrule[\heavyrulewidth]{1-6}
\end{tabular}
\caption{Total training time on various datasets with different number of topics $K$. Entries marked with ``--'' indicates model training is too slow to be finished in 2 days. }
\label{tab:time}
\vspace{-20pt}
\end{table}
Table~\ref{tab:time} compares the total running time of model training with different sized datasets and models. 
As a common practice~\cite{hoffman2013stochastic}, we determine convergence of training when the difference between the test set per-word log-likelihoods of two consecutive iterations is smaller than some threshold.
On small dataset like 20Newsgroups (thousands of documents) and small model (hundreds of topics), all approaches finish training in a reasonable time. However, with increasing number of documents and latent topics, we see that the vanilla CTM model (with $\mathcal{O}(K^3)$ inference complexity) and its scalable version S-CTM (with $\mathcal{O}(K^2)$ inference complexity) quickly becomes impractical, limiting their deployment in real-world scale tasks. Our proposed topic embedding method, by contrast, scales linearly with the topic size, and is capable of handling 100K topics on over 8M documents (PubMed)---a problem size several orders of magnitude larger than previously reported largest results~\cite{chen2013scalable} (1K topics on millions of documents). Notably, even with added model power and increased extraction performance compared to LDA (as has been shown in sections~\ref{sec:cls}-\ref{sec:ir}), our model only imposes negligible additional training time, showing strong potential of our method for practical deployment of real-world large-scale applications as LDA does. 

Figure~\ref{fig:scale}, left panel, shows the convergence curves on NYTimes as training goes. Using similar time, our model converges to a better point (higher test likelihood) than LDA does, while S-CTM is much slower, failing to arrive convergence within the time frame.

\begin{figure}[!h]
\begin{center}
\includegraphics[scale = 0.34]{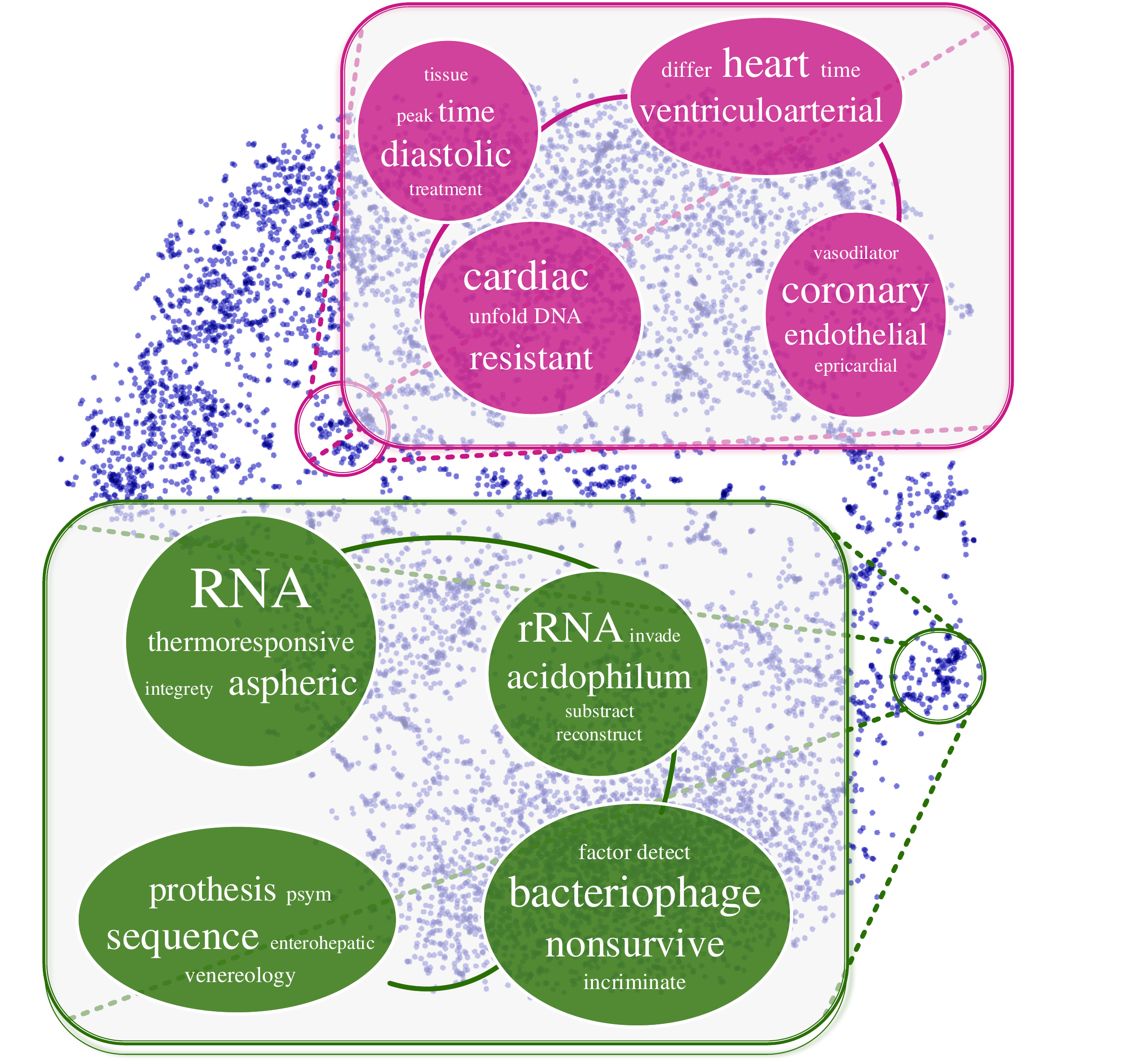}
\vspace{-10pt}
\caption{Visualization of 100K correlated topics on PubMed. See the captions of Figure~\ref{fig:viz-nytimes} for more depictions.}
\label{fig:pubmed}
\end{center}
\vspace{-10pt}
\end{figure}
Figure~\ref{fig:scale}, middle panel, measures the total training time with varying number of topics. We use the small 20Newsgroups dataset since on larger data (e.g., NYTimes and PubMed) the CTM and S-CTM models are usually too slow to converge in a reasonable time. We see that the training time of CTM increases quickly as more topics are used. S-CTM works well in this small data and model scale, but, as have been shown above, it is incapable of tackling larger problems. In contrast, our approach scales as efficiently as the simpler LDA model. Figure~\ref{fig:scale}, right panel, evaluates the runtime of one inference iteration on a minibatch of 500 documents. when the topic size grows to a large number, CTM and S-CTM fail to finish one iteration in 2 hours. Our model, by contrast, keeps as scalable as LDA and considerably speeds up over CTM and S-CTM. 
\subsection{Visualization and Analysis}
We qualitatively evaluate our approach by visualizing and exploring the extracted latent topics and correlation patterns. 

Figure~\ref{fig:viz-20news} visualizes the topic correlation graph inferred from the 20Newsgroups dataset. We can see many topics are strongly correlated to each other and exhibit clear correlation structure. For instance, the set of topics in the right upper region are mainly about astronomy and are interrelated closely, while their connections to information security topics shown in the lower part are weak.
Figure~\ref{fig:pubmed} shows 100K topic embeddings and their correlations on the PubMed dataset. Related topics are close to each other in the embedding space, revealing diverse substructures of themes in the collection.
Our model discovers very meaningful structures, providing insights into the semantics underlying the large text corpora and facilitating understanding of the large collection of topics. 

\section{Conclusions}\label{sec:conclude}
We have developed a new correlated topic model which induces distributed vector representations of latent topics, and characterizes correlations with the closeness of topic vectors in the embedding space. Such modeling scheme, along with the sparsity-aware sampling in inference, enables highly efficient model training with linear time complexity in terms of the model size. Our approach scales to unprecedentedly large data and models, while achieving strong performance in document classification and retrieval. The proposed correlation method is generally applicable to other context, such as modeling word dependencies for improved topical coherence. It is interesting to further speedup of the model inference through variational neural Bayes techniques~\cite{kingma2013auto,goyal2017nonparametric} for amortized variational updates across data examples. Note that our model is particularly suitable to incorporate neural inference networks that, replacing the per-document variational embedding distributions, map documents into compact document embeddings directly. We are also interested in combining generative topic models with advanced deep text generative approaches~\cite{hu2017unifying,hu2017controllable,liang2017recurrent} for improved text modeling.

\section*{Acknowledgments}
This research is supported by NSF~IIS1447676, ONR~N000141410684, and ONR~N000141712463.

\bibliographystyle{ACM-Reference-Format}
\bibliography{sigproc} 

\clearpage

\begin{appendices}

\renewcommand{\theequation}{\Alph{section}.\arabic{equation}}
\onecolumn

\section{Inference}
\subsection{Stochastic Mean-Field Variational Inference}

We first assume a mean-field family of variational distributions:
\begin{equation}
\begin{split}
q(\bm{u},\bm{\phi},\bm{a},\bm{\eta},\bm{z}) = \prod\nolimits_{k}q(\bm{u}_k)q(\bm{\phi}_k) \prod\nolimits_{d}q(\bm{a}_d)q(\bm{\eta}_d)\prod\nolimits_n q(z_{dn}),
\end{split}
\end{equation}
where the factors have the parametric forms:
 \begin{equation}
 \begin{split}
 q(\bm{u}_k)=\mathcal{N}(\bm{u}_k|\bm{\mu}_k, \Sigma^{(u)}_{k}),&\quad q(\bm{a}_d)=\mathcal{N}(\bm{a}_d|\bm{\gamma}_d, \Sigma^{(a)}_{d}),\\
 q(\bm{\phi}_k) = \text{Dir}(\bm{\phi}_k|\bm{\lambda}_k),&\quad q(\bm{\eta}_d) = \mathcal{N}(\bm{\eta}_d|\bm{\xi}_d, \Sigma^{(\eta)}_{d}), \\
 q(z_{dn}) = \text{Multi}(z_{dn}|\bm{\kappa}_{dn})&.
\end{split}
\label{suppeq:vi-factor}
\end{equation}
Variational algorithms aim to minimize KL divergence from $q$ to the true posterior, which is equivalent to tightening the evidence lower bound (ELBO):
\begin{equation}
\begin{split}
\mathcal{L}(q) &= \E_q\left[\log p(\bm{u}, \bm{a}, \bm{\eta}, \bm{z}, \bm{w}, \bm{\phi} | \alpha, \beta, \rho, \tau)\right] - \E_q\left[\log q(\bm{u}, \bm{a}, \bm{\eta}, \bm{z}, \bm{\phi})\right]\\
&= \E_q\left[\log p(\bm{u} | \alpha)\right] + \E_q\left[\log p(\bm{\phi} | \beta)\right] + \mathbb{E}_q\left[\log p(\bm{a} | \rho)\right] + \E_q\left[\log p(\bm{\eta} | \bm{u}, \bm{a}, \tau)\right] \\
& \quad + \mathbb{E}_q\left[\log p(\bm{z} | \bm{\eta})\right] + \mathbb{E}_q\left[\log p(\bm{w} | \bm{\phi}, \bm{z})\right] - \mathbb{E}_q\left[\log q(\bm{u}, \bm{a}, \bm{\eta}, \bm{z}, \bm{\phi})\right].
\end{split}
\end{equation}

\subsection{Optimize $q(\bm{z})$}
\begin{equation}
\label{suppeq:vi-zdn}
\begin{split}
q(z_{dn} = k) &\propto \exp\big\{\mathbb{E}_{-z_{dn}}\left[\log p(z_{dn} =k | \bm{\eta_d})\right] + \mathbb{E}_{-z_{dn}}\left[\log p(w_{dn} | \bm{\phi}_k, z_{dn} = k )\right] \big\} \\
&\propto \exp\big\{\mathbb{E}_{-z_{dn}}\left[\log (\text{softmax}_k(\bm{\eta}_d))\right] + \mathbb{E}_{-z_{dn}}\left[\sum\nolimits_v\bm{1}(w_{dn}=v)\log\phi_{kv}\right] \big\} \\
&\propto \exp\big\{\xi_{dk} + \sum\nolimits_v\bm{1}(w_{dn}=v)(\Psi(\lambda_{kv}) - \Psi(\sum_{v^{\prime}=1}^V\lambda_{kv^{\prime}})) \big\}.
\end{split}
\end{equation}

\textbf{Sparsity-aware topic sampling.} Direct computation of $q(z_{dn})$ with Eq.\eqref{suppeq:vi-zdn} has complexity of $\mathcal{O}(K)$, which becomes prohibitive in the presence of many latent topics. To address this, we exploit two aspects of intrinsic sparsity in the modeling: (1) Though a whole corpus can cover a large diverse set of topics, a single document in the corpus is usually about only a small number of them. We thus only maintain the top $K_s$ entries in each $\bm{\xi}_d$, where $K_s \ll K$, 
making the complexity due to the first term in the right-hand side of Eq.\eqref{suppeq:vi-zdn} only $\mathcal{O}(K_s)$ for all $K$ topics in total; 
(2) A topic is typically characterized by only a few words in the large vocabulary, we thus cut off the variational word weight vector $\bm{\lambda}_k$ for each $k$ by maintaining only its top $V_s$ entries ($V_s \ll V$). Such sparse treatment helps enhance the interpretability of learned topics, 
and allows cheap computation with average $\mathcal{O}(KV_s/V)$ cost for the second term\footnote{In practice we also set a threshold $s$ such that each word $v$ needs to have at least $s$ non-zero entries in $\{\lambda_{k}\}_{k=1}^{K}$. Thus the exact complexity of the second term is $\mathcal{O}(\max\{KV_s/V, s\})$.}.
With the above sparsity-aware updates, the resulting complexity for Eq.\eqref{suppeq:vi-zdn} with $K$ topics is brought down to  $\mathcal{O}(K_s+KV_s/V)$, a great speedup over the original $\mathcal{O}(K)$ cost. The top $K_s$ entries of $\bm{\xi}_d$ are selected using a Min-heap data structure, whose computational cost is amortized across all words in the document, imposing $\mathcal{O}(K/N_d\log K_s)$ computation per word. The cost for finding the top $V_s$ entries of $\bm{\lambda}_k$ is similarly amortized across documents and words, and becomes insignificant. 

Besides, updating the remaining variational parameters will frequently involve computation of variational expectations under $q(z_{dn})$. It is thus crucial to speedup this operation. To this end, we employ sparse approximation by sampling from $q(z_{dn})$ a single indicator $\tilde{z}_{dn}$, and use the ``hard'' sparse distribution $\tilde{q}(z_{dn}=k):=\bm{1}(\tilde{z}_{dn}=k)$ to estimate the expectations. Note that the sampling operation is cheap, having the same complexity with computing $q(z_{dn})$ as above. As shown shortly, such sparse computation will significantly reduce our running cost.

\subsection{Optimize $q(\bm{\phi})$}
For each topic $k$, we isolate only the terms that contain $q(\bm{\phi}_k)$,
\begin{equation}
\label{}
\begin{split}
q(\bm{\phi}_k) &\propto \exp\big\{\mathbb{E}_{-\bm{\phi}_k}(\log \prod\nolimits_v\phi_{kv}^{\beta-1} ) + \mathbb{E}_{-\bm{\phi}_k}(\log\prod\nolimits_{d, n, v}\phi_{kv}^{\bm{1}(w_{dn}=v)\cdot\bm{1}(\tilde{z}_{dn}=k)})\big\}\\
&\propto \prod\nolimits_v\phi_{kv}^{\beta-1+\sum\nolimits_{d,n}\bm{1}(w_{dn}=v)\cdot\bm{1}(\tilde{z}_{dn}=k)}.
\end{split}
\end{equation}
Therefore,
\begin{equation}
\label{ }
q(\bm{\phi}_k) \sim \text{Dir}(\bm{\lambda}_k),
\end{equation}
\begin{equation}
\label{suppeq:vi-lambda}
\lambda_{kv} = \beta + \sum\nolimits_{d,n}\bm{1}(w_{dn}=v)\cdot \bm{1}(\tilde{z}_{dn}=k).
\end{equation}
The cost for updating $q(\bm{\phi})$ is globally amortized across documents and words, and thus insignificant compared with other local parameter update.

\subsection{Optimize $q(\bm{u})$ and $q(\bm{a})$}
\begin{equation}
\label{ }
q(\bm{u}_k) \propto \exp\big\{ \mathbb{E}_{-\bm{u}_k}\left[\log p(\bm{u}_k | \alpha)\right] + \sum\nolimits_d\mathbb{E}_{-\bm{u}_k}\left[\log p(\bm{\eta}_d | \bm{a}_d, \bm{u}, \tau)\right]\big\},
\end{equation}
\begin{equation}
\label{ }
\begin{split}
\mathbb{E}_{-\bm{u}_k}\left[\log p(\bm{u}_k | \alpha)\right] &= \mathbb{E}_{-\bm{u}_k}\left[\log \big\{\frac{1}{(2\pi)^{\frac{M}{2}}\alpha^{-\frac{M}{2}}}\exp(-\frac{\alpha}{2}\bm{u}_k^T\bm{u}_k)\big\}\right]\\
&\propto -\frac{\alpha}{2}\bm{u}_k^T\bm{u}_k,
\end{split}
\end{equation}

\begin{equation}
\label{ }
\begin{split}
\mathbb{E}_{-\bm{u}_k}\left[\log p(\bm{\eta}_d | \bm{a}_d, \bm{u}, \tau)\right] &= \mathbb{E}_{-\bm{u}_k}\left[\log \big\{\frac{1}{(2\pi)^{\frac{M}{2}}\tau^{-\frac{M}{2}}}\exp(-\frac{\tau}{2}(\bm{\eta}_d-\bm{U}\bm{a}_d)^T(\bm{\eta}_d-\bm{U}\bm{a}_d))\big\}\right]\\
&= -\frac{\tau}{2}\bm{u}_k^T\left[\sum\nolimits_d(\Sigma_d^{(a)} + \bm{\gamma}_d\bm{\gamma}_d^T)\right]\bm{u}_k + \tau\sum\nolimits_d\xi_{dk}\bm{\gamma}_d^T\bm{u}_k + C.
\end{split}
\end{equation}
Therefore,
\begin{equation}
\label{u}
q(\bm{u}_k) \propto \exp\big\{ -\frac{1}{2}\bm{u}_k^T\left[\alpha \bm{I} + \sum\nolimits_d(\tau\Sigma^{(a)}_d + \tau\bm{\gamma}_d\bm{\gamma}_d^T)\right]\bm{u}_k + \tau\sum\nolimits_d\xi_{dk}\bm{\gamma}_d^T\bm{u}_k \big\},
\end{equation}
where $\Sigma_d^{(a)}$ is the covariance matrix of $\bm{a}_d$. From Eq.\eqref{u}, we know $q(\bm{u}_k) \sim \mathcal{N}(\bm{\mu}_k, \Sigma_k^{(u)})$.
\begin{equation}
\label{Sigmau}
\Sigma_k^{(u)} = \left[\alpha\bm{I} + \sum\nolimits_d(\tau\Sigma^{(a)}_d + \tau\bm{\gamma}_d\bm{\gamma}_d^T)\right]^{-1}.
\end{equation}
Notice that $\Sigma_k^{(u)}$ is unrelated to $k$, which means all topic embeddings share the same covariance matrix, we denote it as $\Sigma^{(u)}$.
\begin{equation}
\label{suppeq:vi-u}
\bm{\mu}_k = \tau\Sigma^{(u)}\cdot(\sum\nolimits_d\xi_{dk}\bm{\gamma}_d).
\end{equation}
Analogously, 
\begin{equation}
\label{suppeq:vi-a}
\bm{\gamma}_d = \tau\Sigma^{(a)}\cdot(\sum\nolimits_k\xi_{dk}\bm{\mu}_k),
\end{equation}
\begin{equation}
\label{suppeq:vi-Sa}
\Sigma^{(a)} = \left[\gamma\bm{I} + \tau K \Sigma^{(u)} +\sum\nolimits_k\tau\bm{\mu}_k\bm{\mu}_k^T\right]^{-1}.
\end{equation}
Since $\Sigma^{(a)}$ is unrelated to $d$, we can rewrite Eq.\eqref{Sigmau} as
\begin{equation}
\label{suppeq:vi-Su}
\Sigma^{(u)} = \left[\alpha\bm{I} + \tau D\Sigma^{(a)}_d +\sum\nolimits_d \tau\bm{\gamma}_d\bm{\gamma}_d^T\right]^{-1}.
\end{equation}
The cost for optimizing $\bm{\gamma}_d$ is $\mathcal{O}(KDM)$. Updating the set of variational topic vector means $\{\bm{\mu}_k\}_{k=1}^{K}$ and $\Sigma^{(a)}$ both imposes complexity $\mathcal{O}(KM^2)$, and update of $\Sigma^{(u)}$ costs $\mathcal{O}(M^3)$. Since $\bm{\mu}$, $\Sigma^{(a)}$, and $\Sigma^{(u)}$ are all global parameters, we update them in a distributed manner.

\subsection{Optimize $q(\bm{\eta})$}
Assume $q(\bm{\eta}_d)$ is Gaussian Distribution and its covariance matrix is diagonal, i.e., $\eta_{dk} \sim \mathcal{N}(\xi_{dk}, \Sigma_{d}^{(\eta)}), \Sigma_{d}^{(\eta)} = \text{diag}(\bm{\sigma}_d)$.

We can isolate the terms in ELBO including $\bm{\eta}_d$,
\begin{equation}
\label{ }
\mathcal{L}(\bm{\eta}_d) = \mathbb{E}_q\left[\log p(\bm{\eta}_d | \bm{U}, \bm{a}_d)\right] + \mathbb{E}_q\left[\log p(\bm{z}_d | \bm{\eta}_d) \right] - \mathbb{E}_q\left[\log q(\bm{\eta}_d) \right],
\end{equation}

\begin{equation}
\label{ }
\mathbb{E}_q\left[\log p(\bm{\eta}_d | \bm{U}, \bm{a}_d)\right] = -\frac{\tau}{2}\sum\nolimits_k(\xi_{dk}^2 + \sigma_{dk}^2) + \tau\bm{\xi}_d^T\bm{\mu}\bm{\gamma}_d + C,
\end{equation}
\begin{equation}
\label{intract}
\mathbb{E}_q\left[\log p(\bm{z}_d | \bm{\eta}_d) \right] = \sum\nolimits_{k, n}\bm{1}(z_{dn} = k)\mathbb{E}_q\left[\log (\text{softmax}_k(\bm{\eta}_d))\right],
\end{equation}
\begin{equation}
\label{ }
\mathbb{E}_q\left[\log q(\bm{\eta}_d) \right] = -\sum\nolimits_k\log\sigma_{dk} + C.
\end{equation}
For Eq.\eqref{intract}, the expectation is intractable due to normalization term in softmax. As a result, we use reparameterization trick and Monto Carlo estimator to approximate the expectation:
\begin{equation}
\label{ }
\begin{split}
\bm{\eta}^{(t)}_d = \bm{\xi}_d + \bm{\sigma}_d \odot \bm{\epsilon}^{(t)};\quad \bm{\epsilon}^{(t)} \sim \mathcal{N}(\bm{0},\bm{I}),
\end{split}
\end{equation}
where $\odot$ is the element-wise multiplication. With $T$ samples of $\bm{\eta}_d$, we can estimate the variational lower bound and the derivatives $\nabla\mathcal{L}$ w.r.t. the variational parameters $\{\bm{\xi}_d, \bm{\sigma}_d\}$.

\begin{equation}
\label{ }
\nabla_{\bm{\xi}_d}\mathbb{E}_q\left[\log p(\bm{\eta}_d | \bm{U}, \bm{a}_d)\right] = \tau(\tilde{\bm{U}}\bm{\gamma}_d - \bm{\xi}_d).
\end{equation}
Here $\tilde{\bm{U}}$ is the collection of variational means of topic embeddings
where the $k$th row $\tilde{\bm{U}}_{k\cdot}=\bm{\mu}_k$.
\begin{equation}
\label{ }
\begin{split}
\nabla_{\bm{\xi}_d}\mathbb{E}_q\left[\log p(\bm{z}_d | \bm{\eta}_d) \right] &= \mathbb{E}_q\left[\nabla_{\bm{\xi}_d}\log p(\bm{z}_d | \bm{\eta}_d) \right] \\
&\approx \frac{1}{T}\sum\nolimits_{t=1}^T\sum\nolimits_{k, n}\bm{1}(\tilde{z}_{dn} = k)\left[\bm{e}_k - \textbf{softmax}(\bm{\xi}_d + \bm{\sigma}_d \odot \bm{\epsilon}_d^{(l)})\right] \\
&\approx \sum\nolimits_{k,n}\bm{1}(\tilde{z}_{dn}=k) \bm{e}_k - (N_d/T)\sum\nolimits_{t=1}^{T}\textbf{softmax}\left(\bm{\eta}_{d}^{(t)}\right),
\end{split} 
\end{equation}
where $\bm{e}_k$ is an one-hot vector, which evaluates to $1$ in its $k^{th}$ entry. $T$ is the sample number.

\begin{equation}
\label{ }
\nabla_{\bm{\sigma}_d}\mathbb{E}_q\left[\log p(\bm{\eta}_d | \bm{U}, \bm{a}_d)\right] = -\tau\bm{\sigma}_d,
\end{equation}
\begin{equation}
\label{ }
\nabla_{\bm{\sigma}_d}\mathbb{E}_q\left[\log p(\bm{z}_d | \bm{\eta}_d) \right] = \mathbb{E}_q\left[\nabla_{\bm{\sigma}_d}\log p(\bm{z}_d | \bm{\eta}_d) \right] = \bm{0},
\end{equation}
\begin{equation}
\label{suppeq:vi-xi}
\nabla_{\bm{\sigma}_d}\mathbb{E}_q\left[\log q(\bm{\eta}_d) \right] = -\frac{\bm{1}}{\bm{\sigma}_d},
\end{equation}
where $\frac{\bm{1}}{\bm{\sigma}_d}$ is element-wise computation. Therefore,
\begin{equation}
\label{suppeq:vi-xi}
\nabla_{\bm{\xi}_d}\mathcal{L} = \tau(\tilde{\bm{U}}\bm{\gamma}_d - \bm{\xi}_d) + 
\sum\nolimits_{k,n}\bm{1}(\tilde{z}_{dn}=k) \bm{e}_k - (N_d/T)\sum\nolimits_{t=1}^{T}\textbf{softmax}\left(\bm{\eta}_{d}^{(t)}\right),
\end{equation}
\begin{equation}
\label{ }
\nabla_{\bm{\sigma}_d}\mathcal{L} = -\tau\bm{\sigma}_d + \frac{\bm{1}}{\bm{\sigma}_d}.
\end{equation}

We can conclude that $\sigma_{dk} = \tau$ and thus there is no update for $\bm{\sigma}$ in our algorithm. In the experiment, we set $T =1 $ and use \textit{Adagrad} to update $\bm{\xi}_d$. From \eqtref{suppeq:vi-xi}, the time complexity for updating variational mean topic weight vector $\bm{\xi}_d$ is $\mathcal{O}(KM + N_d + K)$.

\end{appendices}

\end{document}